\documentclass{article} %

\usepackage{amsmath,amsfonts,bm}

\def\eqref#1{equation~\ref{#1}}

\def\1{\bm{1}}

\DeclareMathAlphabet{\mathsfit}{\encodingdefault}{\sfdefault}{m}{sl}
\SetMathAlphabet{\mathsfit}{bold}{\encodingdefault}{\sfdefault}{bx}{n}

\usepackage[dvipsnames]{xcolor}
\usepackage{hyperref}
\hypersetup{
    colorlinks=true,
	citecolor=MidnightBlue,
	linkcolor=OrangeRed,
	urlcolor=OrangeRed
}
\usepackage[capitalise,nameinlink]{cleveref} 
\usepackage{iclr2025_conference,times}
\usepackage{url}
\usepackage{booktabs}
\usepackage{multirow}
\usepackage{graphicx}
\usepackage{float}
\usepackage{caption}

\title{PixelGaussian: Generalizable 3D Gaussian Reconstruction from Arbitrary Views}

\author{Xin Fei$^{1,2,}$\thanks{Work done while visiting UC Berkeley.} \quad 
Wenzhao Zheng$^{1,2,}$\thanks{Corresponding author.} \quad 
Yueqi Duan$^1$ \quad 
\textbf{Wei Zhan}$^2$ \quad \\
\textbf{Masayoshi Tomizuka}$^2$ \quad 
\textbf{Kurt Keutzer}$^2$ \quad
\textbf{Jiwen Lu}$^1$ \\
$^1$Tsinghua University \quad 
$^2$University of California, Berkeley\\
\texttt{feix21@mails.tsinghua.edu.cn; wenzhao.zheng@outlook.com}\\
\url{https://wzzheng.net/PixelGaussian}
}

\iclrfinalcopy %
\begin{document}

\maketitle

\begin{figure}[h]
    \centering
    \includegraphics[width=1\linewidth]{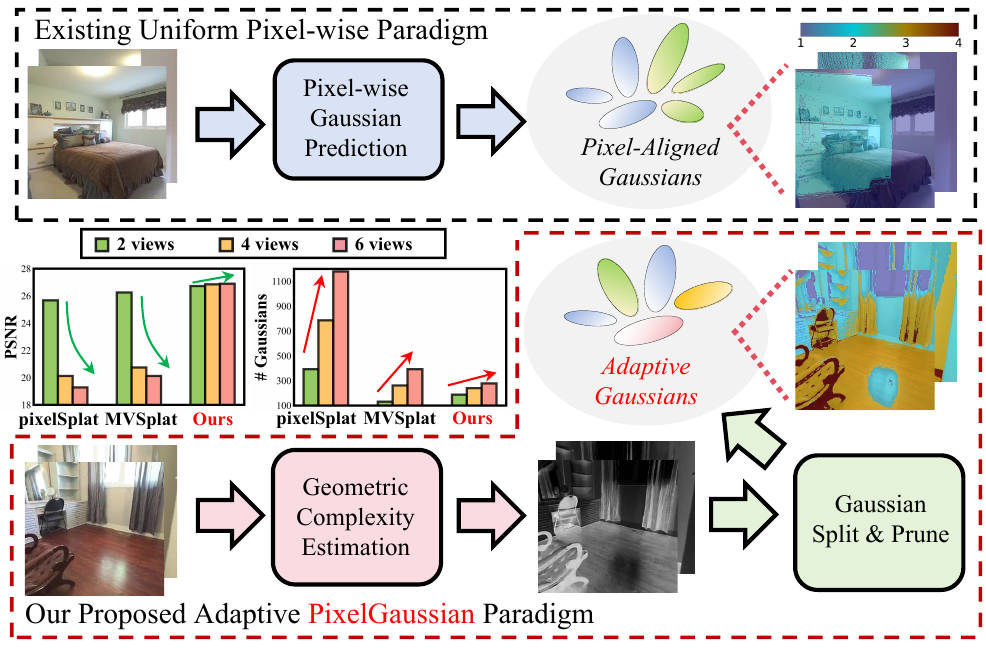}
    \caption{Most existing generalizable 3D Gaussian splatting methods (e.g., pixelSplat~\citep{charatan23pixelsplat}, MVSplat~\citep{chen2024mvsplat}) assign a fixed number of Gaussians to each pixel, leading to inefficiency in capturing local geometry and overlap across views. 
    Differently, our PixelGaussian dynamically adjusts the Gaussian distributions based on geometric complexity in a feed-forward framework.
    With comparable efficiency, PixelGaussian (trained using 2 views) successfully generalizes to various numbers of input views with adaptive Gaussian densities.
    }
    \label{fig:teaser}
    \vspace{0.3cm}
\end{figure}

\begin{abstract}
    We propose \textbf{PixelGaussian}, an efficient feed-forward framework for learning generalizable 3D Gaussian reconstruction from arbitrary views. 
Most existing methods rely on uniform pixel-wise Gaussian representations, which learn a fixed number of 3D Gaussians for each view and cannot generalize well to more input views. 
Differently, our PixelGaussian dynamically adapts both the Gaussian distribution and quantity based on geometric complexity, leading to more efficient representations and significant improvements in reconstruction quality. 
Specifically, we introduce a Cascade Gaussian Adapter to adjust Gaussian distribution according to local geometry complexity identified by a keypoint scorer.
CGA leverages deformable attention in context-aware hypernetworks to guide Gaussian pruning and splitting, ensuring accurate representation in complex regions while reducing redundancy.
Furthermore, we design a transformer-based Iterative Gaussian Refiner module that refines Gaussian representations through direct image-Gaussian interactions. 
Our PixelGaussian can effectively reduce Gaussian redundancy as input views increase.
We conduct extensive experiments on the large-scale ACID and RealEstate10K datasets, where our method achieves state-of-the-art performance with good generalization to various numbers of views. 
Code: \url{https://github.com/Barrybarry-Smith/PixelGaussian}.
\end{abstract}

\section{Introduction}

Novel view synthesis (NVS) seeks to reconstruct a 3D scene from a series of input views and generate high-quality images from previously unseen viewpoints. High-quality and real-time reconstruction and view synthesis are crucial for autonomous driving (\citealp{neurad}; \citealp{khan2024autosplatconstrainedgaussiansplatting}; \citealp{tian2024drivingforwardfeedforward3dgaussian}; \citealp{huang2024textit}), robotics perception (~\citealp{wildersmith2024radiancefieldsroboticteleoperation}; ~\citealp{Jiang_2023}) and virtual or augmented reality (\citealp{yang2024r2humanrealtime3dhuman}; ~\citealp{zheng2024gpsgaussian}). 

NeRF-based methods (~\citealp{mildenhall2020nerf}; ~\citealp{Hu_2022_CVPR}; ~\citealp{liu2020neural}; ~\citealp{neff2021donerf}) have achieved remarkable success by encoding 3D scenes into implicit radiance fields, yet sampling volumes for NeRF rendering is costly in both time and memory. Recently, ~\citet{kerbl3Dgaussians} proposed to represent 3D scenes explicitly using a set of 3D Gaussians, enabling much more efficient and high-quality rendering via a differentiable rasterizer. Still, the original 3D Gaussian Splatting requires separate optimization on each single scene, which significantly reduces inference efficiency.
To tackle this problem, recent researches have aimed at generating 3D Gaussians directly from a feed-forward network without any per-scene optimization (~\citealp{charatan23pixelsplat}; ~\citealp{chen2024mvsplat}; ~\citealp{liu2024mvsgaussian}; ~\citealp{szymanowicz24splatter}; ~\citealp{zheng2024gpsgaussian}). Typically, these approaches adhere to a paradigm where a fixed number of Gaussians is predicted for each pixel in the input views. The Gaussians derived from different views are then directly merged to construct the final 3D scene representation. However, such a paradigm limits the model performance as the Gaussian splats are uniformly distributed across images, making it difficult to capture local geometric details effectively. Additionally, as the number of input views increases, directly merging Gaussians can degrade reconstruction performance due to severe Gaussian overlap and redundancy across views.

To address this, we propose \textbf{PixelGaussian}, which enables dynamic adaption on both 3D Gaussian distribution and quantity. To be specific, we first uniformly initialize Gaussian positions following ~\citet{chen2024mvsplat} to accurately localize the Gaussian centers. To identify geometry complexity across images, we then compute a relevance score map for each input view from image features in an end-to-end manner. Under the guidance of score maps, we construct a Cascade Gaussian Adapter (CGA), which leverages deformable attention~\citep{xia2022visiontransformerdeformableattention} to control the pruning and splitting operations. After CGA, more Gaussians are allocated to regions with complex geometry for precise reconstruction, while unnecessary and duplicate Gaussians across views are pruned to reduce redundancy and improve efficiency. Since these Gaussian representations still fall short in fully capturing the image details, we further introduce a transformer-based Iterative Gaussian Refiner (IGR) to refine 3D Gaussians through direct image-Gaussian interactions. Finally, we employ rasterization-based rendering using the refined Gaussians to generate novel views at target viewpoints.

We conduct extensive experiments on ACID~\citep{infinite_nature_2020} and RealEstate10K~\citep{zhou2018stereomagnificationlearningview} benchmarks for large-scale 3D scene reconstruction and novel view synthesis. 
PixelGaussian outperforms existing methods on different input views with a comparable inference speed. 
Notably, existing generalizable 3D Gaussian splatting methods (pixelSplat~\citep{charatan23pixelsplat} and MVSplat~\citep{chen2024mvsplat}) fail to achieve good results when transferring to more input views while our method demonstrates consistent performance.
This is because existing pixel-wise methods generate uniform pixel-aligned Gaussian predictions, and our model mitigates Gaussian overlap and redundancy across views by dynamically adjusting their distribution based on local geometry complexity.
Visualizations and ablations further demonstrate that both CGA and IGR blocks are crucial in adapting Gaussian distribution, enabling the proposed PixelGaussian to capture geometry details and achieve better reconstruction accuracy.

\section{Related Work}

\textbf{Multi-View Stereo.}
Multi-View Stereo (MVS) aims to reconstruct a 3D representation from multi-view images of a given scene or object. Since accurate depth estimation is essential for reliable 3D reconstruction from 2D inputs, most MVS methods (~\citealp{gu2020cascadecostvolumehighresolution}; ~\citealp{ding2021transmvsnetglobalcontextawaremultiview}; ~\citealp{yao2018mvsnet}) require ground truth depth for supervision in training process. Additionally, point-based MVS approaches generally separate the processes of depth estimation and point cloud fusion processes. Recently, inspired by efficient Gaussian representations proposed by~\citet{kerbl3Dgaussians}, \citet{chen2024mvsplat} introduces to directly predict depth for pixel-wise Gaussians from a cost volume structure without requiring depth supervision, significantly improving model scalability and flexibility. Therefore, following a similar approach, we construct a lightweight cost volume to facilitate depth estimation, which serves as an efficient initialization for 3D Gaussians in our PixelGaussian.

\textbf{Per-scene 3D Reconstruction.} 
Neural Radiance Fields (NeRF) have revolutionized the field of 3D reconstruction by representing scenes as implicit neural fields~\citep{mildenhall2020nerf}. Subsequent researches have focused on overcoming the limitations of the original NeRF to improve its performance and broaden its applicability. Some researches aim to improve the efficiency for novel view synthesis (~\citealp{Hu_2022_CVPR}; ~\citealp{yu2022plenoxels}; ~\citealp{yu2021plenoctrees}; ~\citealp{liu2020neural}; ~\citealp{neff2021donerf}). Moreover, several studies concentrate on capturing intricate geometry and temporal information to achieve accurate and dynamic reconstruction (~\citealp{li2020neural}; ~\citealp{du2021nerflow}; ~\citealp{pumarola2020d}; ~\citealp{23iccv/tian_mononerf}; ~\citealp{Wang_2022_CVPR}). Compared to implicit NeRF-based methods, 3D Gaussian Splatting (3DGS)~\citep{kerbl3Dgaussians} represents a 3D scenario as a set of explicit 3D Gaussians, enabling a rasterization-based splatting rendering process that is significantly more efficient in both time and memory. Given that 3DGS still requires millions of 3D Gaussians to represent a single scene, numerous studies have focused on achieving real-time rendering and minimizing memory usage (~\citealp{fan2023lightgaussian}; ~\citealp{katsumata2024compactdynamic3dgaussian}; ~\citealp{scaffoldgs}). Additionally, some researches focus on enhancing the reconstruction quality of 3DGS by employing multi-scale rendering~\citep{yan2024multiscale3dgaussiansplatting}, advanced shading models~\citep{jiang2023gaussianshader3dgaussiansplatting} or incorporating physically based properties for realistic relighting~\citep{R3DG2023}. However, these methods still require per-scene optimization and rely on dense input views, which can be computationally expensive and limit their scalability for large-scale or dynamic scenes.

\textbf{Generalizable 3D Reconstruction.}
PixelNeRF~\citep{yu2021pixelnerf} pioneers the approach of predicting pixel-wise features directly from input views to reconstruct neural radiance fields. Following methods incorporate volume or transformer architectures to improve the performance of feed-forward NeRF models (~\citealp{chen2021mvsnerf}; ~\citealp{xu2024murf}; ~\citealp{Miyato2024GTA}; ~\citealp{srt22}; ~\citealp{du2023widerender}). However, these feed-forward NeRF approaches typically demand substantial memory and computational resources due to the expensive per-pixel volume sampling process (\citealp{wang2021ibrnet}; \citealp{johari-et-al-2022}; \citealp{barron2021mipnerf}; \citealp{garbin2021fastnerf}; \citealp{reiser2021kilonerfspeedingneuralradiance}; \citealp{mueller2022instant}). With the advent of 3DGS, PixelSplat~\citep{charatan23pixelsplat} initiates a shift towards feed-forward Gaussian-based reconstruction. It takes sparse input views to directly predict pixel-wise 3D Gaussians by leveraging epipolar geometry to learn cross-view features. MVSplat~\citep{chen2024mvsplat} constructs a cost volume structure for depth estimation, which significantly boosts both model efficiency and reconstruction quality. Additionally, MVSGaussian~\citep{liu2024mvsgaussian} further improves model performance by introducing an efficient hybrid Gaussian rendering process. Moreover, SplatterImage~\citep{szymanowicz24splatter} and GPS-Gaussian~\citep{zheng2024gpsgaussian} predict pixel-wise 3D Gaussians for object-centric or human reconstruction.

However, these feed-forward methods are constrained by the pixel-wise Gaussian prediction paradigm, which limits the model's performance as the Gaussian splats are uniformly distributed across images. Such a paradigm inadequately captures intricate geometries, while also causing Gaussian overlap and redundancy across views, ultimately resulting in severe rendering artifacts. In comparison, PixelGaussian consists of a Cascade Gaussian Adapter (CGA), allowing for dynamic adaption on both Gaussian distribution and quantity. Visualizations demonstrate that CGA is capable of allocating more Gaussians in areas rich in geometric details, while reducing duplicate Gaussians in similar regions across input views. Furthermore, we introduce an Iterative Gaussian Refiner (IGR), enabling direct interaction between 3D Gaussians and local image features via deformable attention. Experimental results show that IGR effectively leverages image features to guide Gaussians in capturing the full information contained within the images, significantly enhancing the model's ability to capture local intricate geometry.

\section{Proposed Approach}

In this section, we present our method to learn generalizable Gaussian representations from arbitrary views. Given an arbitrary set of input images $\mathcal{I}= \{I_{i} \}_{i=1}^{N} \in \mathbb{R}^{N \times H \times W \times 3}$ and corresponding camera poses $\mathcal{C}=\{C_{i}\}_{i=1}^{N}$, our PixelGaussian aims to learn a mapping $\mathcal{M}$ from images to 3D Gaussians for scene reconstruction:
\begin{equation}
   \mathcal{M}: \{(I_{i}, C_{i})\}_{i=1}^{N} \mapsto \{(\mu_{j}, s_{j}, r_{j}, \alpha_{j}, sh_{j})\}_{j=1}^{N_K},
\end{equation}
where $N_K$ is the total number of 3D Gaussians, which adaptively varies depending on the scene context. Each Gaussian is parameterized by its position $\mu_{j}$, scaling $s_{j}$, rotation $r_{j}$, opacity $\alpha_{j}$ and spherical harmonics $sh_{j}$.

\begin{figure}[t]
    \centering
    \includegraphics[width=\linewidth]{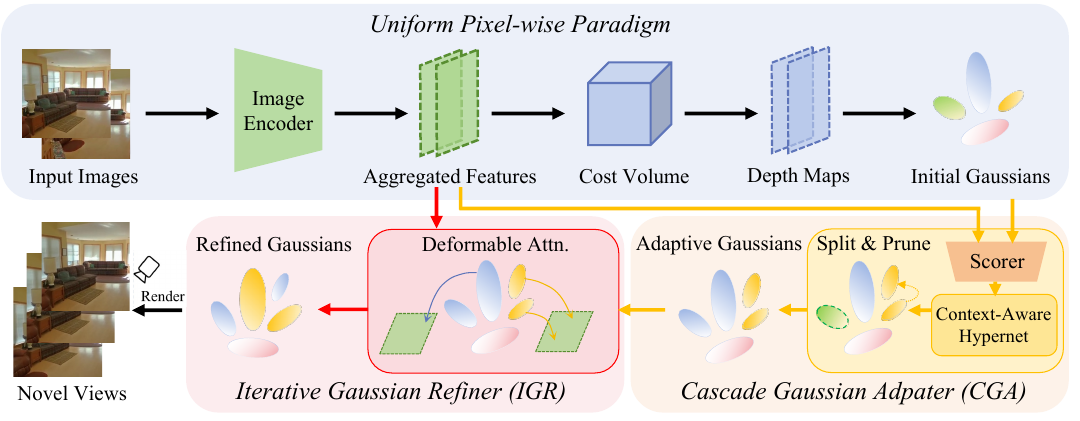}
    \vspace{-0.7cm}
    \caption{\textbf{Overview of PixelGaussian.} Given multi-view input images, we initialize 3D Gaussians using a lightweight image encoder and cost volume. Cascade Gaussian Adapter (CGA) 
    then dynamically adapts both the distribution and quantity of Gaussians. By leveraging local image features, Iterative Gaussian Refiner (IGR) further refines Gaussian representations via deformable attention. Finally, novel views are rendered from the refined 3D Gaussians using rasterization-based rendering.}
    \label{fig:pipeline}
    \vspace{-0.02cm}
\end{figure}

As illustrated in Figure~\ref{fig:pipeline}, we first use a lightweight cost volume for depth estimation and Gaussian position initialization. We then introduce Cascade Gaussian Adapter (CGA), which dynamically adapts both Gaussian distribution and quantity based on local geometric complexity. Finally, we explain how Iterative Gaussian Refiner (IGR) enables direct image-Gaussian interactions, further refining Gaussian distribution and representations for enhanced reconstruction.

\subsection{Gaussian Initialization}
\label{sec:3.1}
\textbf{Position Initialization.}
Following the instructions of MVSplat~\citep{chen2024mvsplat}, we first extract image features via a 2D backbone consisting of CNN and Swin Transformer~\citep{liu2021Swin}. Specifically, CNN encodes multi-view images to corresponding feature maps, while Swin Transformer performs both self-attention and cross-view attention to better leverage information cross views. Then, we obtain the aggregated multi-view features $\mathcal{F}= \{F_{i} \}_{i=1}^{N}$.

To initialize Gaussian positions precisely, we construct a lightweight cost volume~\citep{yao2018mvsnet} for depth estimation, denoted as $\Phi_{depth}$. We then predict Gaussian centers as follows:
\begin{equation}
\label{eq:Equation 1}
    \mu = P^{-1}(\Phi_{depth}(\mathcal{F}),\mathcal{C})
\end{equation}
where $P^{-1}(\cdot)$ stands for unprojection operation.

\textbf{Parameter Initialization.}
For each Gaussian center $\mu_{j}$, we randomly set corresponding scaling $s_{j} \in \mathbb{R}^{3}$, rotation $r_{j} \in \mathbb{R}^{4}$, opacity $\alpha_{j} \in \mathbb{R}^{1}$, spherical harmonics $sh_{j} \in \mathbb{R}^{C}$ within a proper range. we then get the initial Gaussians set $\mathcal{G}=\{(\mu_{j}, s_{j}, r_{j}, \alpha_{j}, sh_{j})\}_{j=1}^{HW} \in \mathbb{R}^{HW \times (11+C)}$.

\subsection{CGA: Cascade Gaussian Adapter}
\label{sec:3.2}
\begin{figure}[t]
    \centering
    \includegraphics[width=\linewidth]{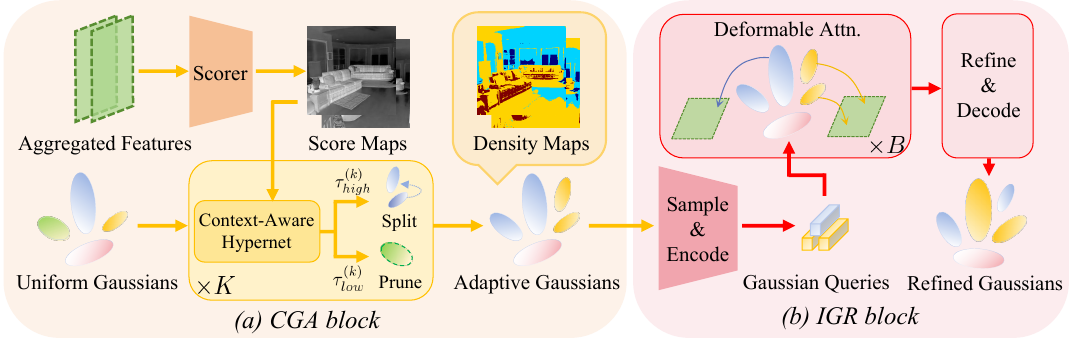}
    \vspace{-0.7cm}
    \caption{\textbf{Illustration of the proposed CGA and IGR Blocks.} (a) CGA comprises a keypoint scorer followed by a series of hypernetworks that produce context-aware thresholds to guide the splitting and pruning of Gaussians. (b) IGR further facilitates direct image-Gaussian interactions, enabling Gaussian representations to capture and extract local geometric features more effectively.}
    \label{fig:block detail}
    \vspace{-0.2cm}
\end{figure}

After obtaining the initial Gaussian set $\mathcal{G}$, we introduce Cascade Gaussian Adapter (CGA) driven by a multi-view keypoint scorer $\Psi$, as shown in Figure~\ref{fig:block detail}(a). CGA contains a set of context-aware hypernetworks $\mathcal{H}$ which dynamically control and guide the following Gaussian pruning and splitting operations. This approach ensures that regions with complex geometry details are represented by a greater number of Gaussians, while areas with poor geometry can be represented with fewer Gaussians. In parallel, CGA effectively removes redundant Gaussians to prevent Gaussian overlap across views. Compared to previous pixel-wise methods, which rigidly allocate a fixed number of Gaussians per pixel, our design dynamically adapts both distribution and quantity of Gaussians based on geometric complexity. This flexibility allows for a more accurate capture of local geometry and mitigates the problem of Gaussian overlap, thereby improving the overall quality of reconstruction.

Given the aggregated features $\mathcal{F}$ derived in Section~\ref{sec:3.1}, $\Psi$ computes relevance score maps $\mathcal{R}=\{R_{i}\}_{i=1}^{N} \in \mathbb{R}^{N \times H \times W}$, where each score map $R_{i}$ is obtained by a learnable weighted average of contributions from different views:

\begin{equation}
\label{eq:Equation 2}
\mathcal{R} = \Psi(\mathcal{F}) = \textit{softmax}\left(\textit{MLP}\left(\sum_{i=1}^{N} \alpha_i \cdot F_i\right)\right), \quad 
\alpha_i = \frac{\exp(\beta_i)}{\sum_{j=1}^{N} \exp(\beta_j)},
\end{equation}

where $A = [\alpha_{1}, \alpha_{2}, \ldots, \alpha_{N}]^T \in \mathbb{R}^{N}$ represents the contribution factor of each view, and is determined by learnable parameters $\beta_{i}(i=1,2,...,N)$.

We first introduce a set of hypernetworks $\mathcal{H}=\{H_{k}\}_{k=1}^{K}$ to generate \textit{context-aware} thresholds. CGA is composed of $K$ stages, where each stage $H_{k}$ takes score maps $\mathcal{R}$ along with Gaussian set $\mathcal{G}_{k}=\{(\mu_{j}^{(k)}, s_{j}^{(k)}, r_{j}^{(k)}, \alpha_{j}^{(k)}, sh_{j}^{(k)})\}_{j=1}^{N_{k}} \in \mathbb{R}^{N_{k} \times (11+C)}$ as input, and outputs thresholds $\tau_{high}^{(k)}, \tau_{low}^{(k)} \in \mathbb{R}$ for splitting and pruning. As illustrated in \eqref{eq:Equation 3}, we first sample and embed Gaussian set $\mathcal{G}_{k}$ into Gaussian score queries $\mathcal{Q}_{r}^{(k)}$. Then we project sampled reference points $\mu^{(k)}$ onto score maps $\mathcal{R}$ with corresponding camera parameters $\mathcal{C}$. Finally, we update queries $\mathcal{Q}_{r}^{(k)}$ with weighted scores from $\mathcal{S}$ and get both thresholds through a simple MLP. Initially, we set $\mathcal{G}_{1}=\mathcal{G}$.

\begin{equation}
\label{eq:Equation 3}
\tau_{high}^{(k)}, \tau_{low}^{(k)} = \mathcal{H}_{k}(\mathcal{G}_{k}, \mathcal{R}, \mathcal{C}) = \textit{MLP}(\sum_{i=1}^{N} \alpha_{i} \cdot \textit{DA}(\mathcal{Q}_{r}^{(k)}, R_{i}, P(\mu^{(k)}, C_{i}))),
\end{equation}

where $DA(\cdot), P(\cdot)$ denote the deformable attention function and projection operation, respectively.

Then, we obtain Gaussian-wise scores by projecting Gaussian centers onto score maps $\mathcal{R}$. To elaborate, let $S_{k} = \{s_{ij}^{(k)}\}\in \mathbb{R}^{N \times N_{k}}$ be the score matrix for Gaussian set $\mathcal{G}_{k}$, where each score $s_{ij}^{(k)}$ is the value at the projection point of the $j-th$ Gaussian center in $R_{i}$, or $0$ if it does not project onto any region in $R_{i}$. The final Gaussian-wise scores $S_{k}^{avg}$ are then computed by averaging scores across different views:

\begin{equation}
\label{eq:Equation 4}
    S_{k}^{avg} = S_{k}^{T} \cdot A,
\end{equation}

Once Gaussian-wise scores are obtained, regions with higher scores, indicating more complex geometry details, undergo splitting operation to allocate more Gaussians for finer representations. For regions with lower scores, we apply an opacity-based pruning operation, gradually reducing Gaussian opacity and scaling to minimize their impact and reduce redundancy.

\textbf{Splitting.} For Gaussian $g_{j}^{(k)} \in \mathcal{G}_{k}$ with score higher than $\tau_{\text{high}}^{(k)}$, we generate \(M\) separate new Gaussians for more detailed representations:

\begin{equation}
\label{eq:Equation 5}
    G_{j}^{(k)}=\textit{SplitNet}(g_{j}^{(k)}) \in \mathbb{R}^{M \times (11+C)},
\end{equation}

where \textit{SplitNet}$(\cdot)$ is a simple MLP-based network that ensures all parameters within proper range. The newly generated Gaussians are then directly concatenated with the existing Gaussian set $\mathcal{G}_{k}$.

\textbf{Pruning.} For Gaussian $g_{j}^{(k)} \in \mathcal{G}_{k}$ with score lower than $\tau_{\text{low}}^{(k)}$, we apply an opacity-based pruning operation rather than directly removing it. Specifically, we set a predefined opacity threshold $\tau_{\alpha}$. If the Gaussian opacity is greater than $\tau_{\alpha}$, we gradually reduce its opacity and scaling:

\begin{equation}
\label{eq:Equation 6}
    \alpha_j^{(k)} \rightarrow \gamma_{\alpha} \cdot \alpha_j^{(k)}, \quad s_j^{(k)} \rightarrow \gamma_{s} \cdot s_j^{(k)},
\end{equation}

where $\gamma_{\alpha} < 1$ and $\gamma_{s} < 1$ are reduction factors. Otherwise, the current Gaussian is removed entirely from Gaussian set $\mathcal{G}_{k}$.

After K-stage adaptation in the Cascade Gaussian Adapter, the initial uniform 3D Gaussian representations are transformed into adaptive forms. Gaussians are redistributed according to geometric complexity, resulting in a more efficient and context-aware representation.

\subsection{IGR: Iterative Gaussian Refiner}

Though CGA allows for a more optimal Gaussian distribution, the Gaussian representations still fall short in capturing the full information contained in the images. Inspired by the efficiency demonstrated by GaussianFormer~\citep{huang2024gaussian} in occupancy prediction, we design a transformer-based Iterative Gaussian Refiner (IGA) to further extract local geometric information from input views, as shown in Figure~\ref{fig:block detail}(b). In this process, we leverage deformable attention to enable direct image-Gaussian interactions, enhancing the ability for 3D Gaussians to more accurately capture intricate geometry details in reconstruction and view synthesis.

IGR is composed of $B$ attention and refinement blocks. In Section~\ref{sec:3.2}, CGA adapts the original Gaussian set $\mathcal{G}$ to $\mathcal{G}=\mathcal{G}_K$. To continue, we first sample and embed $\mathcal{G}$ into Gaussian queries $\mathcal{Q}$. In each block, deformable attention is first applied between Gaussian queries $\mathcal{Q}$ and multi-view features $\mathcal{F}$ to update Gaussian representations. This is followed by a refinement stage where a residual module further fine-tunes the queries. The overall process of IGR can be formulated as:

\begin{equation}
\label{eq:Equation 7}
    \mathcal{Q}_{b} = \Phi_{ref}(\sum_{i=1}^{N} \alpha_{i} \cdot DA(\mathcal{Q}_{b-1}, F_{i}, P(\mu^{(b)}, C_{i}))) \quad b = 1, 2, \dots, B,
\end{equation}

where $DA(\cdot), \Phi_{ref}(\cdot), P(\cdot)$ denote the deformable attention function, refinement layer and projection operation, $F_{i}, C_{i}, \alpha_{i}$ represents the image feature, camera parameters and contribution factor of input view $I_{i}$, respectively. $\mathcal{Q}_{b} (b=1, 2, ..., B)$ stands for output queries from the $b-th$ IGR block, and $\mu^{(b)}$ is the Gaussian center of current stage. Initially, we set $\mathcal{Q}_{0}=\mathcal{Q}$.

Finally, the refined Gaussian queries are decoded into Gaussian parameters $\mathcal{G}_{f}$ through a simple MLP to ensure all parameters within proper range, and then can be used for rasterization-based rendering at novel viewpoints.

\begin{equation}
\label{eq:Equation 8}
    \mathcal{G}_{f}=\{(\mu_{j}^{f}, s_{j}^{f}, r_{j}^{f}, \alpha_{j}^{f}, sh_{j}^{f})\}_{j=1}^{N_{K}}=\textit{MLP}(\mathcal{Q}_{B}).
\end{equation}

Our full model takes ground-truth target RGB images at novel viewpoints as supervision, allowing for efficient end-to-end training. The training loss is calculated as a linear combination of MSE and LPIPS~\citep{zhang2018perceptual} losses, with loss weights of 1 and 0.05, respectively.

Compared to the uniform pixel-wise paradigm, our PixelGaussian approach dynamically adapts both the Gaussian distribution and quantity within the Cascade Gaussian Adapter. Additionally, the Iterative Gaussian Refiner refines Gaussian representations to capture intricate geometric details in the input views. This design achieves more efficient Gaussian distributions while mitigating overlap and redundancy common in pixel-wise methods.

\section{Experiments}

\subsection{Experimental Settings}
\label{sec:Section 4.1}
\textbf{Datasets.} To assess the performance of our model, we conduct experiments on two extensive datasets: ACID~\citep{infinite_nature_2020} and RealEstate10K~\citep{zhou2018stereomagnificationlearningview}. The ACID dataset consists of video frames capturing natural landscape scenes, comprising 11,075 scenes in the training set and 1,972 scenes in the test set. RealEstate10K provides video frames from real estate environments, with 67,477 scenes allocated for training and 7,289 scenes reserved for testing. The model is trained with two reference views, and four novel views are selected for evaluation. During testing, however, we perform multiple experiments where 2, 3, and 4 views are selected for reference, while four novel views selected for evaluation in each scenario.

\textbf{Implementation Details.} We set the resolutions of input images as 256x256. In Cascade Gaussian Adapter (CGA), we apply $K=3$ stages of cascade Gaussian adaption. As for the splitting operation, the SplitNet generates $M=1$ separate new Gaussians, whereas the pruning process uses reduction factors $\gamma_{\alpha}=\gamma_{s}=0.5$ and opacity threshold $\tau_{\alpha}=0.3$. We use $B=3$ blocks in Iterative Gaussian Refiner (IGR) to extract local geometry from input views. We implement our PixelGaussian with Pytorch and train the model on 8 NVIDIA A6000 GPUs for 300,000 iterations with Adam optimizer. More training details are provided in Section~\ref{sec: appendix implementation}.

\subsection{Main Results}

\begin{table}[t]
    \centering
    \caption{\textbf{Results of Novel View Synthesis on the RealEstate10K and ACID benchmarks.} We report the average PSNR and LPIPS~\citep{zhang2018perceptual} on the test set, where all models are trained with 2 reference views and inferred with 2, 3, and 4 reference views.}
    \vspace{-0.3cm}
    \begin{tabular}{cc|cc|cc|cc}
        \toprule
        \multirow{2}{*}{Datasets} & \multirow{2}{*}{Methods} & \multicolumn{2}{c|}{2$\rightarrow$2 Views} & \multicolumn{2}{c|}{2$\rightarrow$3 Views} & \multicolumn{2}{c}{2$\rightarrow$4 Views} \\
        \cmidrule(lr){3-4} \cmidrule(lr){5-6} \cmidrule(lr){7-8}
        &&PSNR$\uparrow$&LPIPS$\downarrow$&PSNR$\uparrow$&LPIPS$\downarrow$&PSNR$\uparrow$&LPIPS$\downarrow$ \\
        \midrule
        \multirow{5}{*}{RealEstate10K} 
        & pixelNeRF & 20.25 & 0.556 & 21.15 & 0.525 & 21.68 & 0.518 \\
        & MuRF & 25.95 & 0.146 & 26.23 & 0.137 & 26.40 & 0.134\\
        & pixelSplat & 25.67 & 0.145 & 22.35 & 0.242 & 20.12 & 0.283  \\
        & MVSplat & 26.25 & 0.130 & 22.94 & 0.236 & 20.74 & 0.268\\
        & PixelGaussian & \textbf{26.72} & \textbf{0.126} & \textbf{26.79} & \textbf{0.123} & \textbf{26.85} & \textbf{0.122}\\
        \midrule
        \multirow{5}{*}{ACID} 
        & pixelNeRF & 20.66 & 0.530 & 21.33 & 0.518 & 21.40 & 0.506 \\
        & MuRF & 27.94 & 0.163 & 28.22 & 0.154 & 28.35 & 0.149 \\
        & pixelSplat & 28.06 & 0.152 & 23.73 & 0.276 & 20.15 & 0.294 \\
        & MVSplat & 28.26 & 0.144 & 23.85 & 0.268 & 20.32 & 0.275\\
        & PixelGaussian & \textbf{28.63} & \textbf{0.140} & \textbf{28.72} & \textbf{0.137} & \textbf{28.78} & \textbf{0.137} \\
        \bottomrule
    \end{tabular}
    \label{tab:main_result}
    \vspace{-0.2cm}
\end{table}

\begin{figure}[t]
    \centering
    \includegraphics[width=\linewidth]{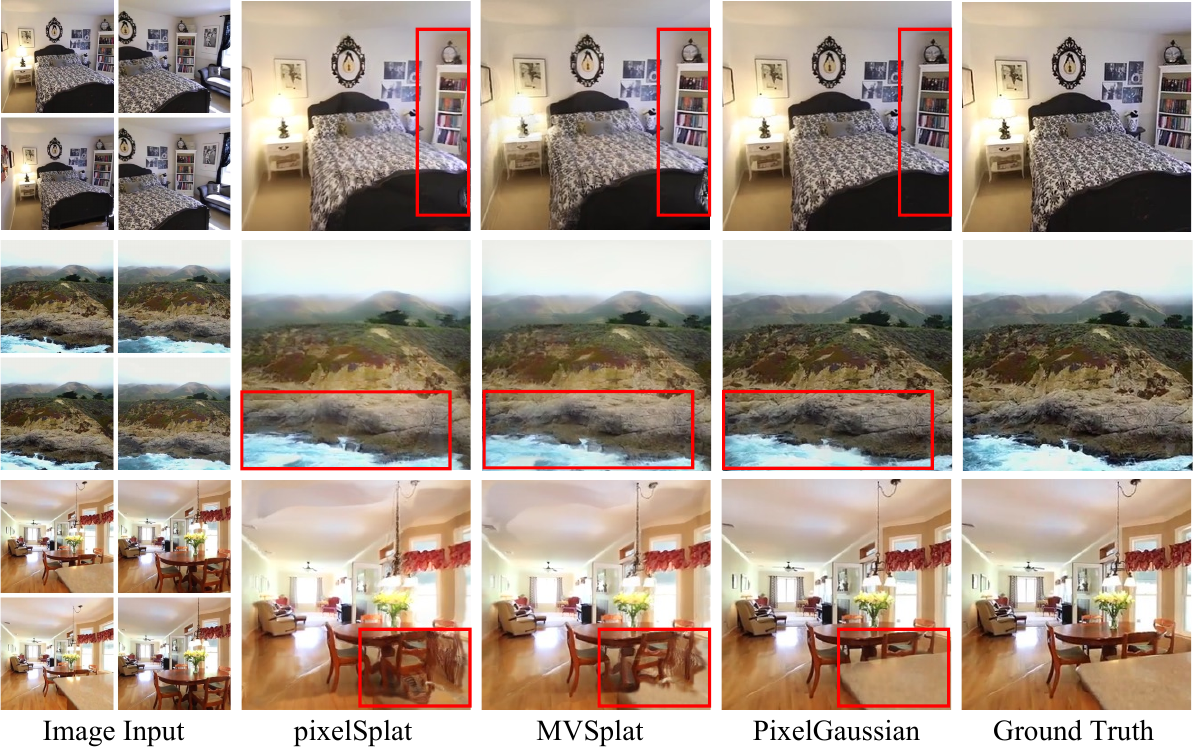}
    \vspace{-0.7cm}
    \caption{\textbf{Visualization results on ACID and RealEstate10K benchmarks.} Pixel-wise methods suffer from Gaussian overlap due to suboptimal Gaussian distributions, whereas PixelGaussian enables dynamic Gaussian adaption and improved local geometry refinement.}
    \label{fig:main_result}
    \vspace{-0.3cm}
\end{figure}

\textbf{Novel View Synthesis.} As shown in Table~\ref{tab:main_result} and Figure~\ref{fig:main_result}, our proposed PixelGaussian consistently outperforms previous NeRF-based methods and pixel-wise Gaussian feed-forward networks across all settings with 2, 3, and 4 reference views. Notably, as the number of input views increases, the reconstruction performance of both pixelSplat~\citep{charatan23pixelsplat} and MVSplat~\citep{chen2024mvsplat} degrades significantly, while PixelGaussian shows a slight improvement. This is because previous methods directly merge multiple views by back-projecting pixel-wise Gaussians to 3D space based on depth maps. Without the capability to adapt the quantity and distribution of Gaussians dynamically, pixel-wise methods often produce duplicated Gaussians with significant overlap, and their spatial positioning is suboptimal. In contrast, PixelGaussian is able to optimize both the distribution and quantity of Gaussians via CGA, while IGR blocks facilitate direct interaction between Gaussian queries and local image features, resulting in more accurate reconstructions. 

\textbf{Multi-View Comparison.} We further compare model performance and Gaussian quantities of different methods across various input views in Table~\ref{tab:PSNR&Gaussians}. Though our method requires more Gaussians than MVSplat~\citep{chen2024mvsplat} with 2 input views due to more frequent splitting than pruning, it achieves better reconstruction with fewer Gaussians as the number of views increases. In regions with richer geometric details, CGA blocks first split more Gaussians for finer representations, followed by IGR to further refine these Gaussians using deformable attention on local image features to better capture and reconstruct geometric details. Meanwhile, CGA prunes duplicate and overlapping Gaussians across views to control the growth of overall Gaussian quantity as the number of input views increases.

\begin{table}[t]
    \centering
    \caption{\textbf{Comparison of PSNR and Gaussian Quantity on RealEstate10K Dataset.} We present the average PSNR and the number of Gaussians (K) for inference using 2, 4, and 6 input views.}
    \vspace{-0.3cm}
    \begin{tabular}{c|cc|cc|cc}
        \toprule
        \multirow{2}{*}{Methods} & \multicolumn{2}{c|}{2$\rightarrow$2 Views} & \multicolumn{2}{c|}{2$\rightarrow$4 Views} & \multicolumn{2}{c}{2$\rightarrow$6 Views} \\
        \cmidrule(lr){2-3} \cmidrule(lr){4-5} \cmidrule(lr){6-7}
        &PSNR$\uparrow$&\# Gaussians&PSNR$\uparrow$&\# Gaussians&PSNR$\uparrow$&\# Gaussians\\
        \midrule
        pixelSplat & 25.67 & 393 K & 20.12 & 786 K & 19.36 & 1179 K \\
        MVSplat & 26.25 & \textbf{131 K} & 20.74 & 262 K & 20.24 & 393 K \\
        PixelGaussian & \textbf{26.72} & 188 K & \textbf{26.85} & \textbf{240 K} & \textbf{26.89} & \textbf{278 K} \\
        \bottomrule
    \end{tabular}
    \label{tab:PSNR&Gaussians}
    \vspace{-0.2cm}
\end{table}

\textbf{Efficiency Analysis.} We explore the efficiency of PixelGaussian compared with dominant pixel-wise methods on a single NVIDIA A6000 GPU. All models are inferred with 4 input views on the RealEstate10K~\citep{zhou2018stereomagnificationlearningview} dataset. We report the average inference latency, memory cost, number of Gaussians and rendering FPS in Table~\ref{tab:efficiency}. Undeniably, PixelGaussian requires higher latency and memory usage than MVSplat~\citep{chen2024mvsplat} due to the extra cost of CGA and IGR blocks. However, this trade-off allows PixelGaussian to achieve higher rendering FPS by utilizing fewer Gaussians and far more better reconstruction quality as the number of input view increases. 

\begin{table}[t]
    \centering
    \caption{\textbf{Efficiency analysis.} Inference on RealEstate10K~\citep{zhou2018stereomagnificationlearningview} dataset using 4 input views, reporting average latency, memory cost, Gaussian quantity, and rendering FPS.}
    \vspace{-0.3cm}
    \begin{tabular}{c|c|c|c|c}
        \toprule
        Methods & Latency & Memory & \# Gaussians & Rendering FPS \\
        \midrule
        pixelSplat & 298 ms & 11.78 G & 786 K & 64  \\
        MVSplat & \textbf{126 ms} & \textbf{3.17 G} & 262 K & 133 \\
        PixelGaussian & 235 ms & 4.39 G & \textbf{240 K} & \textbf{140} \\
        \bottomrule
    \end{tabular}
    \label{tab:efficiency}
    \vspace{-0.2cm}
\end{table}

\subsection{Expeimental Analysis}
In this section, we further investigate and conduct experiments to demonstrate the effectiveness of our PixelGaussian. We first visualize the score maps $\mathcal{S}$ and Gaussian density maps. Then, we present the cascade adaption process of CGA. Finally, we conduct ablation studies on our model. These experiments show that CGA dynamically adapts both the distribution and quantity of Gaussians according to geometric complexity, while IGR further extract local features via direct image-Gaussian interactions, offering significant improvements over traditional pixel-wise methods.

\begin{figure}[t]
    \centering
    \includegraphics[width=\linewidth]{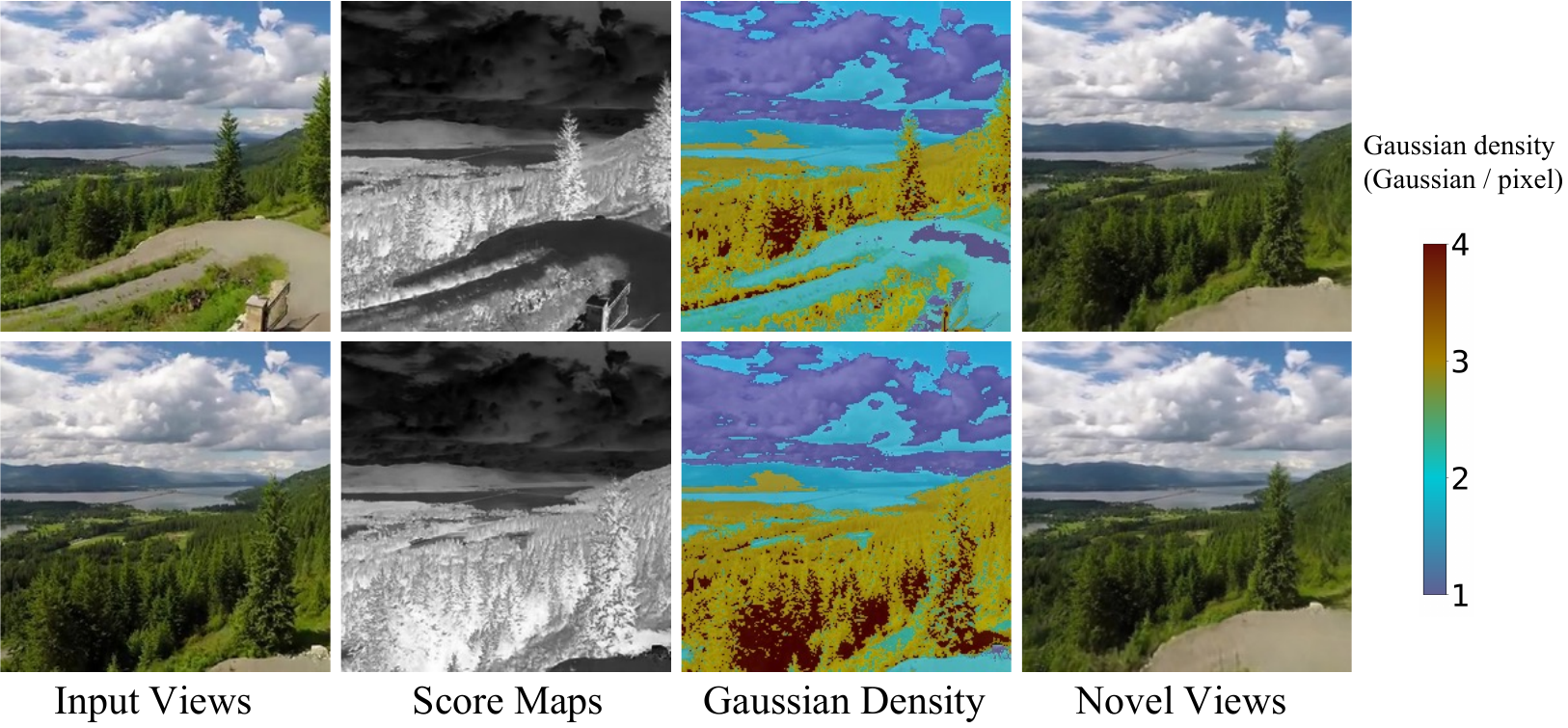}
    \vspace{-0.7cm}
    \caption{\textbf{Visualization of score maps and Gaussian distributions on RealEstate10K dataset.} Cascade Gaussian Adapter dynamically adjusts Gaussian distribution and quantity based on score maps. More Gaussians are allocated to detail-rich regions for more precise representations, while pruning minimizes Gaussian redundancy and overlap across views.}
    \label{fig:CGA_visual}
    \vspace{-0.2cm}
\end{figure}

\textbf{Score Maps and Density Maps.} As shown in \ref{eq:Equation 2}, relevance score maps $\mathcal{S}$ are derived from multi-view image features through the keypoint scorer $\Psi$. To further understand the significance of $\mathcal{S}$ and its impact on the subsequent Gaussian splitting and pruning operations, we visualize the relevance score maps on the RealEstate10K~\citep{zhou2018stereomagnificationlearningview} dataset. Furthermore, we project the centers of the adaptive Gaussian after CGA $(i.e. \{\mu_{j}^{f}\}_{j=1}^{N_K})$, onto each input view as Gaussian density maps. Figure~\ref{fig:CGA_visual} illustrates that in this end-to-end learning framework, keypoint scorer $\Psi$ is able to learn such score maps where local regions with richer and more complex geometric details tend to receive higher score, which guides the following SplitNet to allocate more Gaussians for more precise representations. In contrast, regions with fewer geometric details receive lower scores, leading to pruning to reduce representation complexity while preserving overall representation efficiency.

\begin{figure}[t]
    \centering
    \vspace{0.05cm}
    \includegraphics[width=\linewidth]{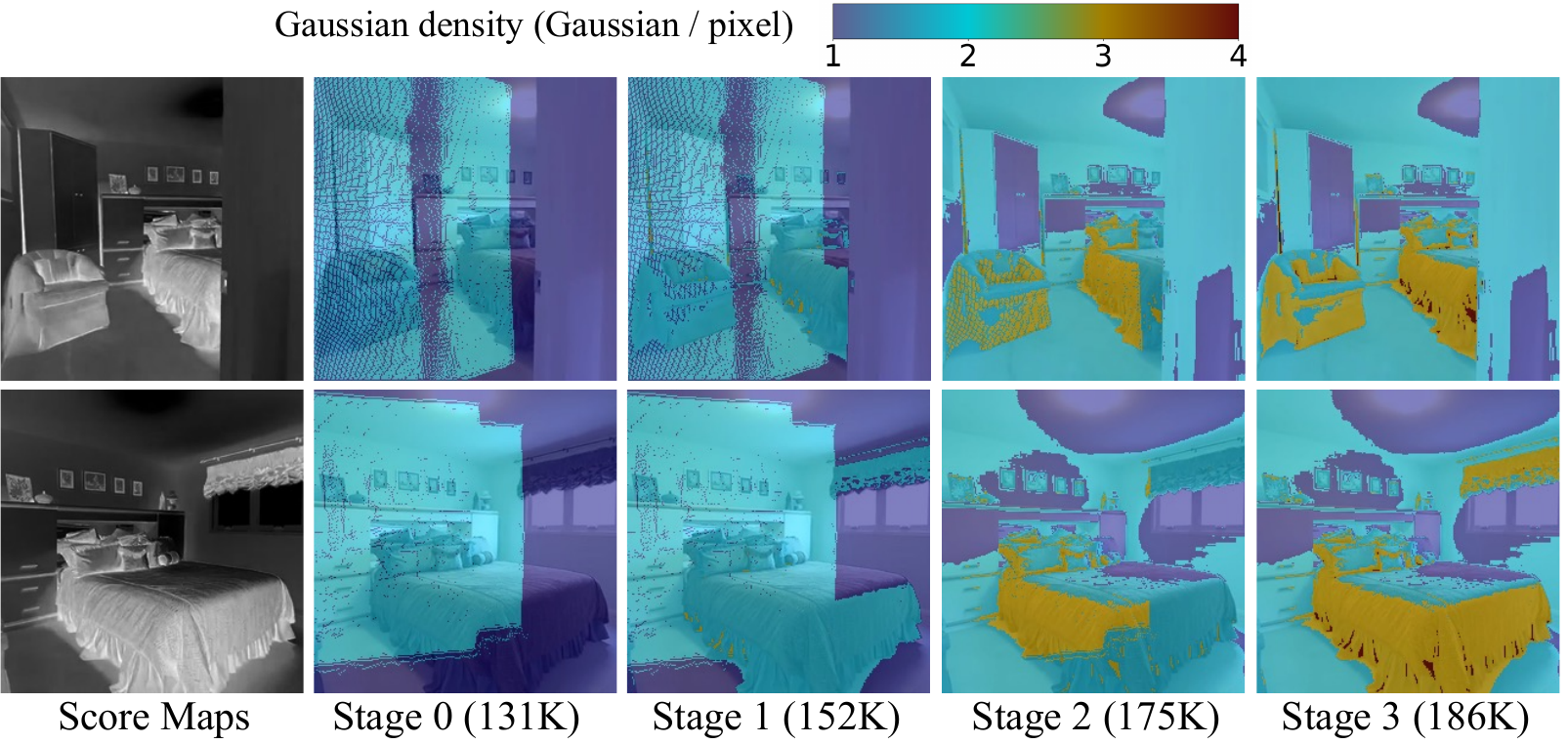}
    \vspace{-0.7cm}
    \caption{\textbf{Visualization of Gaussian adaption process in Cascade Gaussian Adapter.} During inference on RealEstate10K~\citep{zhou2018stereomagnificationlearningview} dataset, CGA progressively concentrates more Gaussians on geometrically complex regions, while pruning duplicate Gaussians across views to control the growth of overall Gaussian quantity.}
    \label{fig:CGA_adapt_process}
    \vspace{-0.1cm}
\end{figure}

\textbf{Cascade Gaussian Adaption.} To present the cascade Gaussian adaption process in CGA blocks more explicitly, we visualize Gaussian density maps during model inference on the RealEstate10K~\citep{zhou2018stereomagnificationlearningview} dataset with 2 input views. As illustrated in Figure~\ref{fig:CGA_adapt_process}, the original Gaussian set $\mathcal{G}$ is initialized on a pixel-wise basis. Following this, score maps $\mathcal{S}$ guide hypernetworks $\mathcal{H}$ to generate context-aware thresholds, which in turn direct the subsequent Gaussian splitting and pruning operations. More Gaussians are allocated in regions with richer geometric details, while duplicate Gaussians across views tend to be pruned, leading to a more efficient and optimal Gaussian distribution for scene representation.

\textbf{Ablation Study.} To further investigate the architecture of PixelGaussian, we conduct ablation studies by inferring our model on RealEstate10K~\citep{zhou2018stereomagnificationlearningview} test dataset with 4 input views. We first introduce a vanilla model, where the initial Gaussian set $\mathcal{G}$ is directly used to render novel views. Then, we adopt rigid CGA blocks without context-aware Hypernetworks $\mathcal{H}$, which means Gaussian set $\mathcal{G}$ goes through splitting and pruning based on fixed thresholds $(\tau_{high}^{(k)}=0.8, \tau_{low}^{(k)}=0.2, k=1,2,...,K)$. We further add HyperNetworks $\mathcal{H}$ to generate context-aware thresholds. Finally, we adopt IGR blocks to refine the Gaussian representations via image-Gaussian interactions. As shown in Table~\ref{tab:ablation}, HyperNetworks $\mathcal{H}$ utilizes score maps $\mathcal{S}$ to generate context-aware thresholds, enabling a more dynamic and efficient Gaussian distribution for scene representation compared to rigid splitting and pruning. Furthermore, IGR blocks refine the Gaussian set iteratively via deformable attention between Gaussians and image features, enhancing their ability to describe and reconstruct intricate local geometric details.

\begin{table}[t]
    \centering
    \caption{\textbf{Ablations on the components of PixelGaussian.} We report the average PSNR, LPIPS~\citep{zhang2018perceptual}, and the number of Gaussians (K) of model inference. }
    \vspace{-0.3cm}
    \begin{tabular}{c|c|c|c}
        \toprule
Methods&PSNR$\uparrow$&LPIPS$\downarrow$&\#Gaussians\\
        \midrule
        Vanilla & 20.34 & 0.272 & 262 K \\
        + Rigid Cascade Gaussian Adapter & 22.46 & 0.220 & 225 K \\
        + HyperNetworks $\mathcal{H}$ & 25.80 & 0.140 & \textbf{240 K} \\
        + Iterative Gaussian Refiner & \textbf{26.85} & \textbf{0.122} & \textbf{240 K} \\
        \bottomrule
    \end{tabular}
    \label{tab:ablation}
    \vspace{-0.3cm}
\end{table}

\section{Conclusion and Discussions}

In this paper, we have presented PixelGaussian to learn generalizable 3D Gaussian reconstruction from arbitrary input views. The core innovation of our approach is \textit{context-aware} Cascade Gaussian Adapter (CGA), which dynamically splits Gaussians in regions with complex geometric details and prunes redundant ones. Further, we incorporate deformable attention within Iterative Gaussian Refiner (IGR), facilitating direct image-Gaussian interactions to improve local geometry reconstructions. Compared to previous uniform pixel-wise methods, PixelGaussian is able to dynamically adapt both Gaussian distribution and quantity guided by the complexity of local geometry details, allocating more to detailed regions and reducing redundancy across views, thus leading to better performance in reconstruction and view synthesis.

\textbf{Discussions and Limitations.} Although PixelGaussian can adjust the distribution of 3D Gaussians dynamically, the initial Gaussians are still derived from pixel-wise unprojection. When we initialize the Gaussian centers completely at random, the model fails to converge. Moreover, deformable attention in IGR consumes substantial computational resources when the number of Gaussians is extremely large, highlighting the need for a more efficient approach to represent 3D scenes with fewer Gaussians. Furthermore, PixelGaussian is unable to perceive the unseen parts of 3D scenes beyond the input views, suggesting the potential need to incorporate generative models.

\clearpage

\appendix
\section{Appendix}

\subsection{Preliminary}
\label{sec: preliminary}
\subsubsection{3D Gaussian Splatting}
3D Gaussian Splatting~\citep{kerbl3Dgaussians} represents a 3D scene as a set of explicit Gaussian primitives as follows:
\begin{equation}
\label{eq:Equation 9}
    \mathcal{G}=\{g_{i}|g_{i}=(\mu_{i}, \Sigma_{i}, \alpha_{i}, sh_{i})\}_{i=i}^{N}
\end{equation}
where each Gaussian has a center $\mu_{i}$, a covariance $\Sigma_{i}$, an opacity $\alpha_{i}$ and spherical harmonics $sh_{i}$. Furthermore, given the scaling matrix $S$ and rotation matrix $R$, we can calculate the covariance matrix:
\begin{equation}
\label{eq:Equation 10}
    \Sigma = RSS^{T}R^{T}
\end{equation}
As explicit Gaussian primitives $\mathcal{G}$ can be rendered via an rasterization-based operation, such approach is much more cheaper in both time and memory compared to implicit neural fields (~\citealp{mildenhall2020nerf}; ~\citealp{barron2021mipnerf}; ~\citealp{23iccv/tian_mononerf}; ~\citealp{chen2021mvsnerf}; ~\citealp{garbin2021fastnerf}) or voxel-based representations(~\citealp{sitzmann2019deepvoxels}; ~\citealp{ji2017surfacenet}; ~\citealp{xie2019pix2vox}; ~\citealp{ogn2017}; ~\citealp{choy20163dr2n2unifiedapproachsingle}). 

\subsubsection{Deformable Attention}
Since the introduction of ViT~\citep{dosovitskiy2021imageworth16x16words}, numerous efficient attention mechanisms have been proposed to further enhance the scalability and reduce the complexity of ViT (~\citealp{liu2021Swin}; ~\citealp{chen2021regionvit};~\citealp{wang2021pyramidvisiontransformerversatile};~\citealp{yang2021focal};~\citealp{zhang2021multiscalevisionlongformernew};~\citealp{dong2022cswintransformergeneralvision};~\citealp{jaegle2021perceivergeneralperceptioniterative};~\citealp{sun2022visualparserrepresentingpartwhole};~\citealp{zhu2020deformable};~\citealp{yue2021visiontransformerprogressivesampling};~\citealp{Chen_2021}). Among them, Deformable Attention Transformer (DAT)~\citep{xia2022visiontransformerdeformableattention} stands out for its ability to effectively capture multi-scale features and adaptively focus on important regions while maintaining lightweight computational overheads. Therefore, we apply deformable attention in both the Score Hypernetworks $\mathcal{H}$ (\eqref{eq:Equation 3}) and Iterative Gaussian Refiner (IGR) (\eqref{eq:Equation 7}) in our PixelGaussian model. We elaborate \eqref{eq:Equation 7} as an example.

Given Gaussian centers $\mu$, corresponding queries $\mathcal{Q}$, feature maps $\mathcal{F}=\{F_{i}\}_{i=1}^{N}$ and camera parameters $\mathcal{C}=\{C_{i}\}_{i=1}^{N}$, we first project Gaussian centers to pixel coordinates to get reference points, denote as $\mathcal{R}=\{R_{i}\}_{i=1}^{N}$:
\begin{equation}
\label{eq:Equation 11}
    R_{i} = P(\mu, C_{i}), (i=1,2,...,N)
\end{equation}
where $P(\cdot)$ denotes projection operation. Next, for Gaussian queries $\mathcal{Q}$, we perform deformable sampling from feature maps $\mathcal{F}$ at reference points $\mathcal{R}$ using bilinear interpolation as follows:
\begin{equation}
\label{eq:Equation 12}
    \phi(F_{i}, (r_x, r_y)) = \sum_{(p_x, p_y)} g(r_x, p_x) \cdot g(r_y, p_y)\cdot  F_{i}(p_x, p_y)
\end{equation}
where $g(x, y)=max(0, 1-|a-b|), (r_x, r_y), F_{i}$ denote reference point and feature map, respectively. Then we can update queries $\mathcal{Q}$ via attention:
\begin{align}
\label{eq:Equation 13}
\mathcal{Q} = \sum_{i=1}^{N} \alpha_{i} \cdot (W_{i} \cdot \phi(F_{i}, R_{i})) \\
W_{i} = softmax(\mathcal{Q} \cdot \phi(F_{i}, R_{i})^T)
\end{align}
where $\alpha_{i}, (i=1,2,...,N)$ represents view weights as illustrated in \eqref{eq:Equation 2}. Following this paradigm, we incorporate deformable attention in the score hypernetworks $\mathcal{H}$ to generate context-aware thresholds, and in IGR blocks to further refine Gaussian representations using image features.

\subsection{More Implementation Details}
\label{sec: appendix implementation}
\textbf{Training Details.} As mentioned in Section~\ref{sec:Section 4.1}, our model is trained on 8 NVIDIA A6000 GPUs, The batch size for a single GPU is set to 4, where each batch contains a large-scale 3D scene with two reference views and four inference views. Detailed settings of image backbone to generate aggregated features $\mathcal{F}$, context-aware score hypernetworks $\mathcal{H}$ in Cascade Gaussian Adapter and Iterative Gaussian Refiner are illustrated in Table~\ref{tab:network architecture}. For MLP-based networks in \eqref{eq:Equation 5} and \eqref{eq:Equation 8}, they all follow criteria in Table~\ref{tab:Gaussian Parameters} to ensure all Gaussian parameters within proper range. 

\begin{table}[h]
\centering
\captionsetup{skip=5pt}

\begin{minipage}[t]{0.5\linewidth}
    \centering
    \caption{Details of Gaussian Parameter Range.}
    \begin{tabular}{c|c}
        \toprule
        Gaussian Parameters & Criterion \\
        \midrule
        scaling $s \in \mathbb{R}^3$ & $s_x, s_y, s_z \in [0.50, 15.00]$ \\
        rotation $r \in \mathbb{R}^4$ & $||r|| = 1$ \\
        opacity $\alpha \in \mathbb{R}$ & $\alpha \in [0, 1]$ \\
        harmonics $sh \in \mathbb{R}^{C}$ & decaying coefficients \\
        \bottomrule
    \end{tabular}
    \label{tab:Gaussian Parameters}
\end{minipage}\hfill
\begin{minipage}[t]{0.5\linewidth}
    \centering
    \caption{Details of Training Settings}
    \begin{tabular}{c|c}
        \toprule
        Config & Setting \\
        \midrule
        optimizer & Adam \\
        scheduler & Cosine Annealing \\
        learning rate & $2 \times 10^{-4}$\\
        weight decay & $1 \times 10^{-4}$ \\
        \bottomrule
    \end{tabular}
    \label{tab:Training Settings}
\end{minipage}

\vspace{0.5cm} 

\caption{Details of Network Architecture.}
\begin{tabular}{c|c}
    \toprule
    Block & Setting \\
    \midrule
    CNN layers & [4, 4, 4] \\
    CNN channels & [32, 64, 128] \\
    Transformer layers & [2, 2, 2, 2, 2, 2] \\
    Transformer channels & [128, 128, 128, 128, 128, 128] \\
    Score HyperNetworks layers & [2, 2, 2] \\
    Score HyperNetworks channels & [1, 1, 1] \\
    IGR layers & [2, 2, 2] \\
    IGR channels & [128, 128, 128]\\
    \bottomrule
\end{tabular}
\label{tab:network architecture}
\end{table}

\textbf{Network Architecture.} As illustrated in Table~\ref{tab:network architecture}, the image backbone comprises a ResNet-based CNN~\citep{he2015deepresiduallearningimage} for feature extraction and a Swin Transformer~\citep{liu2021Swin} for multi-view attention. For context-aware score hypernetworks $\mathcal{H}$ in CGA, we employ $K=3$ stages for adaptive Gaussian adaption. Then, we adopt $B=3$ deformable attention blocks in IGR for Gaussian representation refinement.

\subsection{More Results}
\label{sec: appendix results}
\begin{figure}[h]
    \centering
    \includegraphics[width=\linewidth]{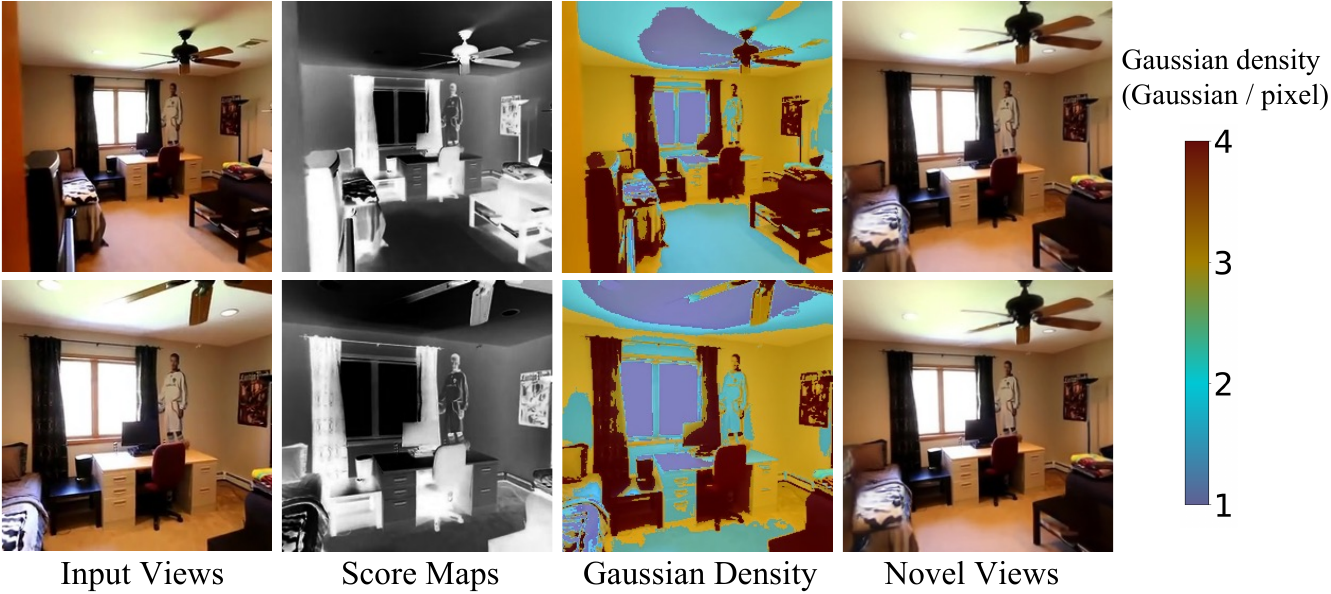}
    \caption{Visualization of score maps and Gaussian distributions of Cascade Gaussian Adapter.}
    \label{fig:CGA appendix}
\end{figure}

\begin{figure}[h]
    \centering
    \includegraphics[width=\linewidth]{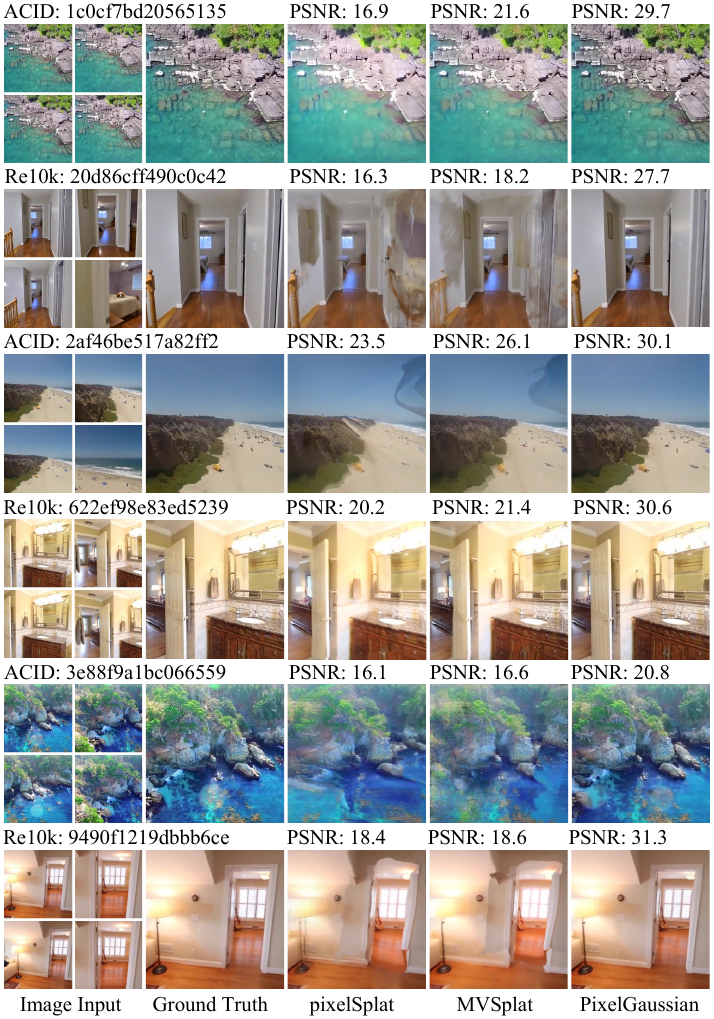}
    \caption{Visualization results on ACID~\citep{infinite_nature_2020} and RealEstate10K~\citep{zhou2018stereomagnificationlearningview} benchmarks.}
    \label{fig:visual appendix}
\end{figure}

We provide more visualization comparisons of PixelGaussian and previous dominant pixel-wise methods (\citealp{charatan23pixelsplat}; \citealp{chen2024mvsplat}) on ACID~\citep{infinite_nature_2020} and RealEstate10K~\citep{zhou2018stereomagnificationlearningview} benchmarks. Additionally, we provide visualizations of more score maps and Gaussian density maps from our PixelGaussian model. As shown in Figure~\ref{fig:CGA appendix} and Figure~\ref{fig:visual appendix}, PixelGaussian benefits from dynamic adaption of Gaussian distributions, resulting in superior reconstruction compared to previous pixel-wise methods.

\end{document}